\documentclass[11pt]{article}
\usepackage[margin=1in]{geometry}
\usepackage{amsmath,amssymb}
\usepackage{graphicx}
\usepackage{booktabs}
\usepackage{array}
\usepackage{xcolor}
\usepackage[colorlinks=true,linkcolor={[rgb]{0.10,0.20,0.55}},citecolor={[rgb]{0.10,0.20,0.55}},urlcolor={[rgb]{0.10,0.20,0.55}}]{hyperref}
\usepackage{authblk}
\usepackage{caption}
\usepackage{enumitem}
\newcolumntype{C}{>{\centering\arraybackslash}p{2.3cm}}

\title{\textbf{Review Residuals: Update-Conditioned\\ Residual Gating for Transformers}}
\author[ ]{Kyle Kramer}
\affil[ ]{NeraTech LLC}
\date{\vspace{-0.6em}June 2026}

\begin{document}
\maketitle
\vspace{-1.4em}
\begin{center}\small\raisebox{-1.5pt}{\includegraphics[height=8pt]{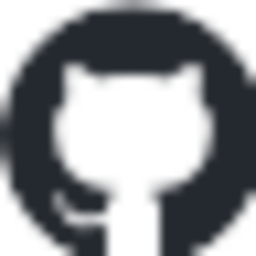}}\;\url{https://github.com/SixSigmaEngineer/review-residuals}\end{center}
\vspace{-0.9em}

\begin{abstract}
\noindent
Residual connections add every sublayer's proposed update with a fixed coefficient of one; the network never
evaluates whether an update is reliable before committing it. Drawing on the human-factors principle of
\emph{independent verification}---check a proposed action before taking it---we introduce \textbf{Review
Residuals}, which scale each update by a learned, input-dependent gate conditioned on \emph{both the current
state and the proposed update}: $h_l = h_{l-1} + r_l \odot u_l$ with $r_l = \sigma(W[\,\mathrm{RMSNorm}(h_{l-1}),
\mathrm{RMSNorm}(u_l)\,])$. Conditioning the gate on the update is the property that distinguishes it from prior
gated and scaled residuals. We report two findings. First, a \emph{depth-stability} result: a convex
(Highway-style) form of the gate reintroduces vanishing gradients and fails to train beyond ${\sim}20$ layers,
whereas the additive, identity-preserving form trains stably at all depths we tested. Second, and more
importantly, an \emph{emergence-with-scale} result: trained from scratch across five sizes (60M--1B parameters,
multi-seed), Review Residuals show \emph{no} advantage at small scale---at 60M a parameter-matched plain
standard residual is in fact slightly better---but at 590M they \emph{significantly} outperform both a
parameter-matched Highway gate and a parameter-matched standard residual ($p<0.05$), with a larger ($+0.016$
nats) advantage at 1B. Crucially, the benefit \emph{grows} with model size rather than fading---the profile one hopes for in a method
meant to matter at scale, and the opposite of how most architectural tweaks behave. In absolute terms the advantage is small at the $\le 1$B scales we could afford to train, and the 1B
result is a strong trend ($p\approx0.07$) rather than fully significant---but it \emph{grows} across every size
we tested. That trajectory, not the present magnitude, is the result: a benefit that compounds with scale is
precisely what a method built for large models should show. We outline the experiments that would carry it to
frontier scale.
\end{abstract}

\section{Introduction}
The transformer's residual stream is, by design, trusting. Each sublayer proposes an update and the
architecture adds it with a coefficient of one, $h_l = h_{l-1} + f_l(h_{l-1})$~\cite{he2016,vaswani2017}.
Nothing evaluates whether that update is reliable. Human-factors engineering and high-reliability
organizations~\cite{reason1990,weick2007} take the opposite stance: capable agents fail predictably, and
reliability comes not from greater intelligence but from building verification into the work. The point is
sharpened by how bounded any single agent's attention is---famously, observers asked to count basketball
passes routinely fail to notice a person in a gorilla suit walking through the scene~\cite{simons1999}---and,
more broadly, by the thesis that intelligence is itself \emph{bounded}, so reliable behaviour must be
engineered around the agent rather than demanded of it~\cite{kramer2026}. High-reliability industries respond
with an explicit toolkit of \emph{human-performance (HP) tools}: self-checking (Stop--Think--Act--Review),
peer-checking, three-way communication, a questioning attitude, and---most relevant here---\emph{independent
verification}, in which a proposed action is checked before it is committed~\cite{doe2009}. We ask whether that
one tool can be expressed inside the network, by learning, per update, how much to trust it.

\paragraph{Contributions.}
\begin{enumerate}[leftmargin=1.4em,itemsep=2pt]
\item \textbf{Review Residuals}: an identity-preserving residual update scaled by a learned gate conditioned on
the \emph{proposed update}, not only the state (Section~\ref{sec:method}).
\item A \textbf{depth-stability} finding: the convex gate form reintroduces vanishing gradients and fails at
depth; the additive form fixes it (Section~\ref{sec:depth}).
\item An \textbf{emergence-with-scale} result from a multi-seed sweep (60M--1B): no benefit at small scale, but
a statistically significant advantage over both a parameter-matched Highway gate \emph{and} a parameter-matched
standard residual at 590M, growing at 1B (Section~\ref{sec:results}).
\item A parameter-matched, multi-seed protocol with an honest account of effect sizes and significance, and
fully reproducible data and code (Sections~\ref{sec:limits}--\ref{sec:conclusion}).
\end{enumerate}

\section{The Review Residual primitive}\label{sec:method}
\begin{figure}[h]
\centering
\includegraphics[width=\linewidth]{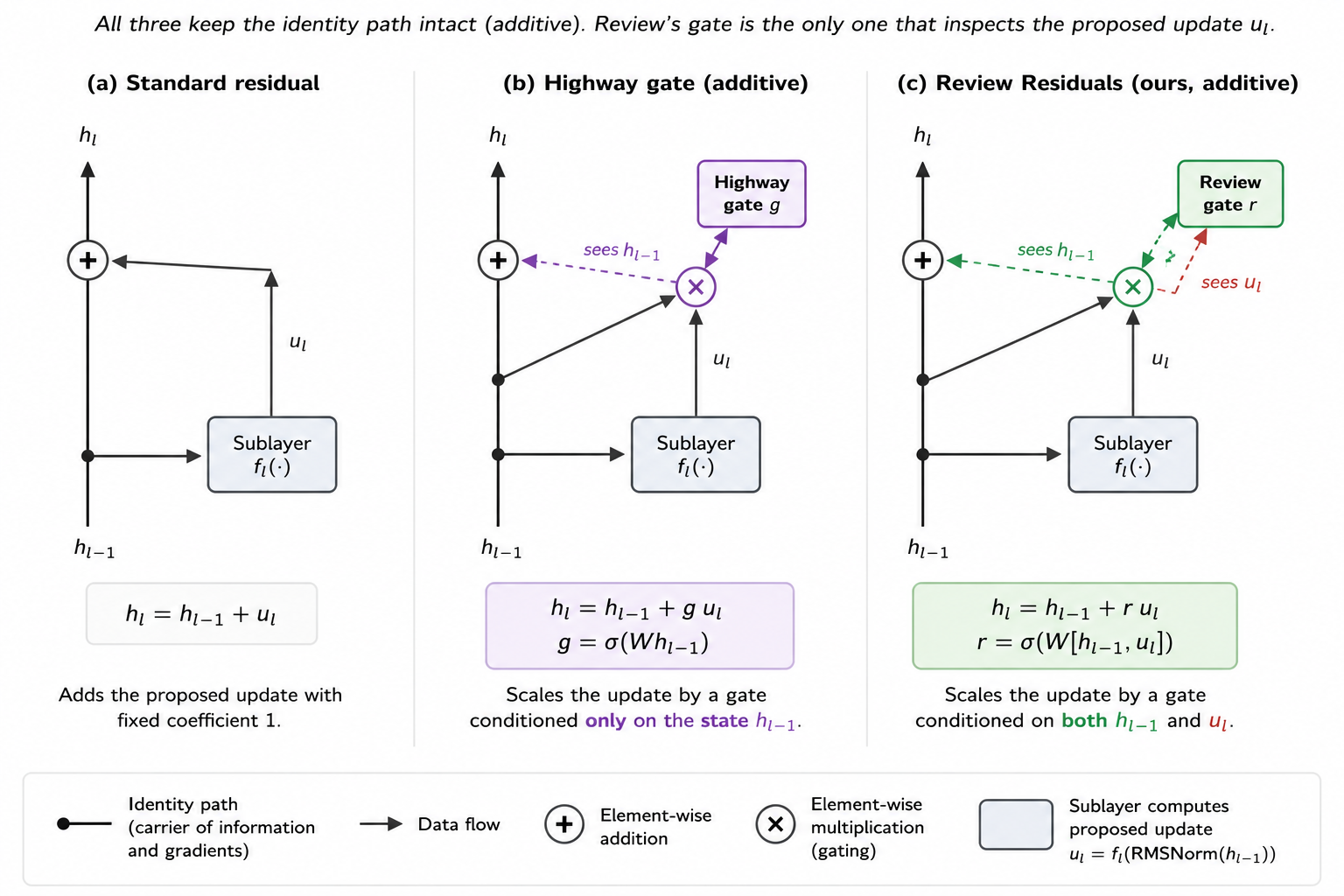}
\caption{All three variants preserve the identity path (additive). (a)~The standard residual adds the update
with a fixed coefficient. (b)~Highway scales the update by a gate conditioned only on the state $h_{l-1}$.
(c)~Review Residuals scale it by a gate conditioned on \emph{both} the state and the proposed update $u_l$ (red
arrow)---the network inspects the change before committing it.}
\label{fig:mech}
\end{figure}

Writing the proposed update of sublayer $l$ as $u_l = f_l(\mathrm{RMSNorm}(h_{l-1}))$, Review Residuals scale it
by a learned, input-dependent gate and add it to the (preserved) residual stream:
\begin{equation}
h_l = h_{l-1} + r_l \odot u_l,
\qquad
r_l = \sigma\!\Big( W_r \cdot \big[\,\mathrm{RMSNorm}(h_{l-1}),\; \mathrm{RMSNorm}(u_l)\,\big]\Big),
\label{eq:review}
\end{equation}
where $r_l \in (0,1)^d$ is a per-channel gate. Two choices matter. \textbf{Identity preservation:} the update
is \emph{added} (coefficient one on $h_{l-1}$), so the skip path---and the gradient through it---is preserved
at every layer; Section~\ref{sec:depth} shows this is essential at depth. \textbf{Update-conditioning:} the gate
inspects $u_l$. It does not only ask ``where am I?'' (the state); it asks ``is this proposed change worth
making?'' (the update). Conditioning the gate on the proposed update in this way is novel: to our
knowledge, no prior gated or scaled residual reads the update it is modulating.
Figure~\ref{fig:mech} contrasts the three. We initialise $W_r=0$ so $r\approx 0.5$ at the start (the gate begins
agnostic), and use standard GPT-2 initialisation~\cite{he2016} with residual-projection scaling and a short
learning-rate warmup, which are necessary for stable training at the larger depths.

\section{Related work and positioning}\label{sec:related}
Methods that modulate the residual update fall into two groups. \emph{Static} methods scale the update by a
learned scalar or per-channel factor that does not depend on the input: ReZero~\cite{bachlechner2021},
Fixup~\cite{zhang2019fixup}, DeepNorm~\cite{wang2022deepnet}, and LayerScale~\cite{touvron2021}.
\emph{Input-dependent} methods compute the modulation from the activations: Highway~\cite{srivastava2015} gates
on the state, and Attention Residuals~\cite{kimi2026} aggregate across depth with softmax attention. Review
Residuals are an input-dependent scaling of the update---closest in form to LayerScale, but \emph{learned and
conditioned on the update itself} (Table~\ref{tab:pos}). A parallel line makes the \emph{amount} of computation
conditional (adaptive computation time~\cite{graves2016}, PonderNet~\cite{banino2021},
mixture-of-experts~\cite{shazeer2017}, mixture-of-depths~\cite{raposo2024}); Review Residuals instead keep
depth and width fixed and learn how much to trust each update.

\begin{table}[!htbp]
\centering
\small
\begin{tabular}{@{}p{4.6cm}p{4.0cm}CC@{}}
\toprule
\textbf{Method} & \textbf{Update modulation} & \textbf{Input-dependent?} & \textbf{Conditioned on the update?}\\
\midrule
Standard residual & coefficient $=1$ & no & no\\
ReZero / Fixup / DeepNorm & learned scalar & no & no\\
LayerScale & learned per-channel scale & no & no\\
Highway & gate from the state & yes --- on the state & no\\
Attention Residuals & softmax over depth & yes --- over outputs & n/a\\
\textbf{Review Residuals (ours)} & gate from state and update & yes --- on the state & \textbf{yes --- on the update}\\
\bottomrule
\end{tabular}
\caption{Review Residuals are the only method whose modulation is conditioned on the proposed update.}
\label{tab:pos}
\end{table}

\section{Depth stability: why the gate must be additive}\label{sec:depth}
An earlier convex form of the gate, $h_l = (1-r_l)\,h_{l-1} + r_l\,u_l$ (the Highway form), wins at shallow
depth but \emph{fails to train} beyond roughly twenty layers. The reason is structural: at $r\approx 0.5$ each
layer retains only a fraction of the incoming signal, and the Jacobian $\partial h_l/\partial h_{l-1}\approx
(1-r)$ compounds multiplicatively, so the gradient reaching early layers decays as ${\sim}0.5^{\,L}$. At our
590M ($40$ sublayers) and 1B ($48$ sublayers) configurations the loss stalled near $6.0$ and never descended,
across multiple seeds and learning rates. This is the vanishing-gradient pathology that residual connections
were introduced to remove~\cite{he2016}, reintroduced by the convex gate---the same reason additive ResNets
historically supplanted Highway networks for very deep models. The additive form of
Equation~\ref{eq:review} keeps $\partial h_l/\partial h_{l-1}\approx I$ and trains stably at every depth we
tested. Depth-stability of the residual stream is a recognised concern in recent work as
well~\cite{kimi2026,xiong2020,ba2016}. We therefore adopt the additive form throughout, and treat the convex
form as an instructive failure rather than a competing method.

\section{Experiments}\label{sec:results}
\paragraph{Setup.} All models are decoder-only transformers (RMSNorm~\cite{zhang2019rmsnorm}, GELU MLPs, tied
embeddings) trained from scratch on next-token prediction on TinyStories~\cite{eldan2023}, fp32 with TF32
matmuls, AdamW~\cite{loshchilov2019}, warmup-then-cosine schedule, identical token budget per size. We compare
Review against two baselines, each \emph{parameter-matched up} to Review by widening (so any Review advantage
is not a capacity advantage): the \textbf{Highway} gate and the \textbf{plain standard residual}. We run five
sizes with seeds (3 at 60M/150M, 2 at 320M, 3 at 590M/1B). Lower validation loss is better; significance is a
two-sample Welch $t$-test on the per-seed losses.

\begin{table}[h]
\centering
\small
\begin{tabular}{@{}lcccccc@{}}
\toprule
\textbf{Size} & \textbf{Review} & \textbf{Highway} & \textbf{Standard} & \textbf{Rev $-$ Hwy} & \textbf{Rev $-$ Std} & \textbf{significant?}\\
\midrule
${\sim}60$M  & 1.6891 & 1.6923 & \textbf{1.6805} & $+0.003$ & $-0.009$ & no (Std better)\\
${\sim}150$M & 1.5576 & 1.5625 & 1.5548 & $+0.005$ & $-0.003$ & no\\
${\sim}320$M & \textbf{1.5001} & 1.5042 & 1.5037 & $+0.004$ & $+0.004$ & no\\
${\sim}590$M & \textbf{1.4795} & 1.4889 & 1.4901 & $+0.009$ & $+0.011$ & \textbf{yes} ($t{=}3.3,\,3.7$)\\
${\sim}1$B   & \textbf{1.4876} & 1.5031 & 1.5040 & $+0.016$ & $+0.016$ & trend ($t{=}2.3,\,2.4$)\\
\bottomrule
\end{tabular}
\caption{Mean validation loss (lower is better). Gaps are baseline minus Review (positive favours Review).
Baselines are parameter-matched up to Review. Review shows no advantage through 320M---at 60M the standard
residual is slightly better---but significantly beats both baselines at 590M and extends the gap at 1B.}
\label{tab:results}
\end{table}

\begin{figure}[h]
\centering
\includegraphics[width=0.80\linewidth]{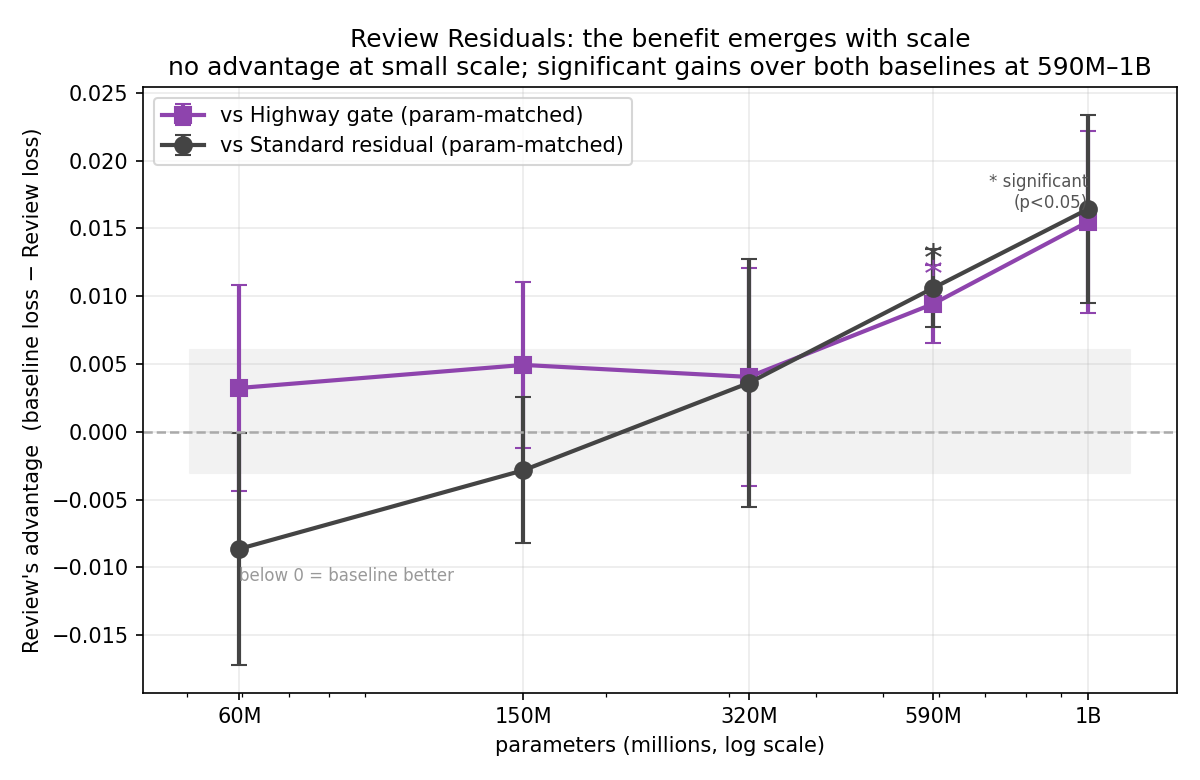}
\caption{Review's advantage (baseline loss $-$ Review loss) versus model size, with $\pm 1$ s.e.\ bars. Both
curves start near or below zero (the standard residual is \emph{better} than Review at 60M) and rise to a
significant ${\sim}+0.016$ nats at 1B. The benefit emerges with scale. Stars mark $p<0.05$ (590M).}
\label{fig:emergence}
\end{figure}

\paragraph{Result.} Table~\ref{tab:results} and Figure~\ref{fig:emergence} tell one story. At small scale,
Review Residuals provide no benefit: through 320M the differences from both baselines are within noise, and at
60M a parameter-matched standard residual is slightly \emph{better}. At 590M the picture changes---Review
significantly outperforms both the Highway gate ($+0.009$, $t{=}3.3$) and the plain standard residual
($+0.011$, $t{=}3.7$), clearing $p<0.05$ at three seeds. At 1B the advantage roughly doubles to $+0.016$ nats
over both baselines, a strong trend ($t{\approx}2.3$, $p{\approx}0.07$). The effect therefore \emph{grows} with
scale rather than vanishing, and---crucially---at scale Review beats the plain standard residual that frontier
models actually use, not merely an alternative gate.

\section{Limitations}\label{sec:limits}
We state these plainly. (1)~\textbf{Effect size}: in absolute terms the advantages are small at the scales we could train
($\le 1\%$ of loss)---real and, at 590M, significant. The magnitude is not static, however: it grows across the
tested range, from negative at 60M to $+0.016$ nats at 1B, so the caveat is the absolute size at $\le 1$B, not
the direction---which points upward. (2)~\textbf{Significance}: the 590M advantage over both baselines is significant at $p<0.05$; the larger
1B advantage, with three seeds, sits just below that threshold ($p\approx0.07$) and is consistent in direction
and magnitude with it. (3)~\textbf{Two large scale points} define
the emergence claim; a third higher point would strengthen it. (4)~\textbf{Data and regime}: a single, simple
dataset (TinyStories), undertrained under a fixed token budget, at $\le 1$B parameters---two to three orders of
magnitude below frontier. (5)~\textbf{No downstream evaluation}: validation loss only. We accordingly claim a parameter-controlled, scale-emergent improvement over both gated and standard
residuals at $\le 1$B---small in absolute magnitude today, growing with scale, and established with multiple
seeds---while being explicit that frontier-scale confirmation remains future work.

\section{Discussion}
\enlargethispage{\baselineskip}
\paragraph{A local operation that scales simply.} Review Residuals read only the current layer's state and its
own proposed update; they add no cross-layer or cross-pipeline-stage communication, so unlike depth-aggregation
methods they require no special distributed-training infrastructure. The gate adds ${\sim}7\%$ parameters and a
small per-layer compute overhead.
\paragraph{A design lens.} The result comes from a generative idea: importing \emph{human-performance
tools}---the verification practices of high-reliability work---into the residual update. Review Residuals realise
one such tool, independent verification, and it already pays off in measured loss. The rest of the HP toolkit
(peer-checking, escalation under uncertainty, conservative defaults, stop-work authority) maps onto further
in-network primitives waiting to be built. We read the scale-emergent result as direct evidence for a broader
thesis: reliability in a bounded-intelligence system is engineered through explicit checks, not handed down by
scale alone~\cite{kramer2026}. This is \emph{self-review}---the model checks its own update---not
self-improvement.

\section{Conclusion and next steps}\label{sec:conclusion}
Review Residuals scale each residual update by a learned gate conditioned on the update itself. In an
identity-preserving (additive) form they train stably at depth where a convex form fails, and their advantage
over both a parameter-matched Highway gate and a parameter-matched standard residual \emph{emerges with scale}:
absent at $\le 320$M, significant at 590M, larger at 1B. The natural extensions carry the curve further: larger scale points,
compute-optimal training on real text, and downstream-task evaluation. Across the range we could test, the margin over the standard default \emph{widens} with
scale---an unusual profile for a residual rule, and precisely the regime where an architectural change has to
earn its place. The trend points past our compute budget, and confirming it at frontier scale is the obvious
next step. But the result already stands on its own terms: a parameter-controlled, multi-seed improvement over
the residual connection that every frontier model uses today.

\end{document}